\documentclass[letterpaper]{article} % DO NOT CHANGE THIS
\usepackage{aaai24}  % DO NOT CHANGE THIS
\usepackage{times}  % DO NOT CHANGE THIS
\usepackage{helvet}  % DO NOT CHANGE THIS
\usepackage{courier}  % DO NOT CHANGE THIS
\usepackage[hyphens]{url}  % DO NOT CHANGE THIS
\usepackage{graphicx} % DO NOT CHANGE THIS
\usepackage{natbib}  % DO NOT CHANGE THIS AND DO NOT ADD ANY OPTIONS TO IT
\usepackage{caption} % DO NOT CHANGE THIS AND DO NOT ADD ANY OPTIONS TO IT
\usepackage{algorithm}
\usepackage{algorithmic}

\usepackage{amsmath,amssymb,amsfonts}
\usepackage{subcaption}
\usepackage{subfig}
\usepackage{textcomp}
\usepackage{xcolor}
\usepackage{defs-custom}
\usepackage{booktabs}
\usepackage{tabularx}
\usepackage{multirow}
\usepackage{adjustbox}
\usepackage{newfloat}
\usepackage{listings}
\DeclareCaptionStyle{ruled}{labelfont=normalfont,labelsep=colon,strut=off} % DO NOT CHANGE THIS
\floatstyle{ruled}
\newfloat{listing}{tb}{lst}{}
\floatname{listing}{Listing}
\title{Guided Game Level Repair via Explainable AI}
\author {
    Mahsa Bazzaz,
    Seth Cooper
}
\affiliations {
    Khoury College of Computer Sciences, Northeastern University\\
    bazzaz.ma@northeastern.edu, se.cooper@northeastern.edu
}

\begin{document}
\maketitle

\begin{abstract}
Procedurally generated levels created by machine learning models can be unsolvable without further editing. Various methods have been developed to automatically repair these levels by enforcing hard constraints during the post-processing step. However, as levels increase in size, these constraint-based repairs become increasingly slow. This paper proposes using explainability methods to identify specific regions of a level that contribute to its unsolvability. By assigning higher weights to these regions, constraint-based solvers can prioritize these problematic areas, enabling more efficient repairs. Our results, tested across three games, demonstrate that this approach can help to repair procedurally generated levels faster.
\end{abstract}

\section{Introduction}
\begin{figure*}[t]
  \centering
  \includegraphics[width=1\textwidth]{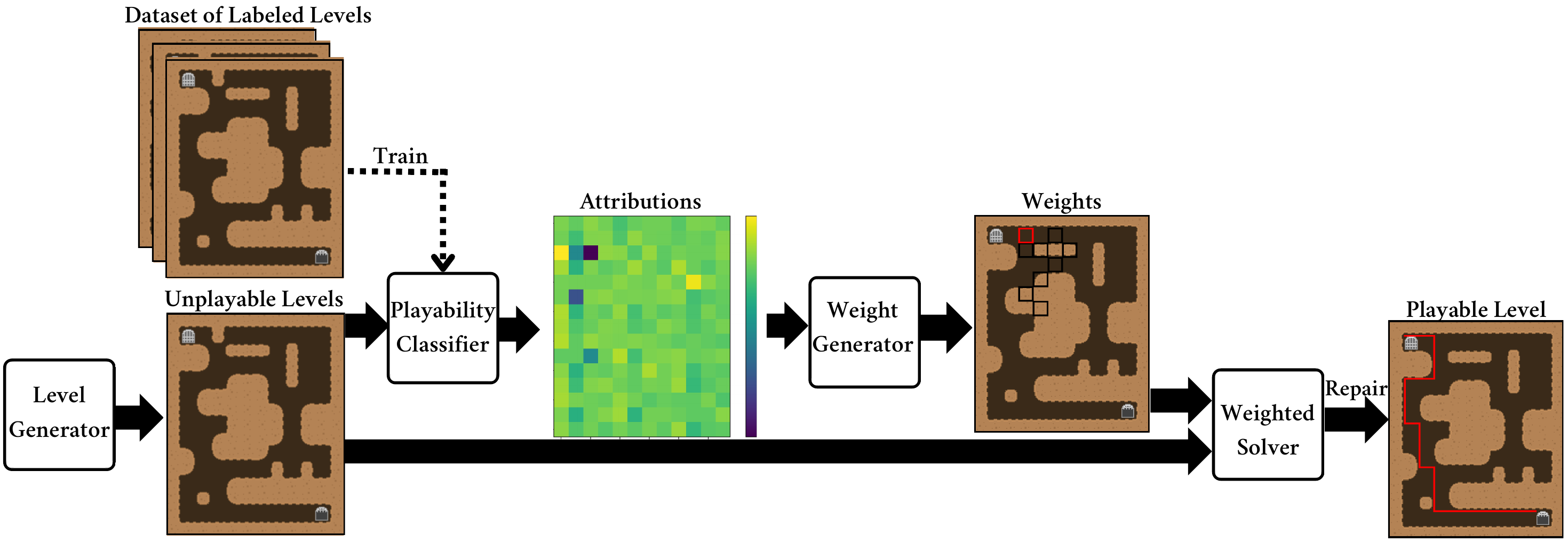}
  \caption{Visual Representation of the system pipeline.}
  \label{fig:system_overview}
\end{figure*}
% state of the world ...
Procedural Content Generation via Machine Learning (PCGML) involves generating game content using models trained on existing game content \citep{summerville2018procedural}. These models aim to produce solvable levels, which players can complete by moving from the start to the end of the level through a series of valid movements while completing tasks and avoiding obstacles.

However, due to the noisy nature of ML-based generation, a repair step is often necessary in PCGML approaches to make the generated levels usable \citep{chen2020image}. The initial outputs from these models often include levels that are impossible to complete or contain broken structures with missing or incorrectly arranged elements (e.g., pipes in platformer games or broken decorations in puzzle games). Consequently, recent research has focused on developing methods to repair these levels by making the necessary adjustments.
% \todo{emphesis on what percentage of levels are unsolvable}

% current limitations ...
One popular approach to level repair involves using constraint satisfaction. This method represents the problem as a collection of constraints over variables and seeks an assignment of values to these variables that satisfies all constraints. Constraint solvers are software tools that search through possible assignments to find one that satisfies the constraints or they prove that no such assignment exists.

In the context of level repair, the process generally starts with the largest possible set of candidate values for each tile. The solver then cycles between propagating the constraints and narrowing down the possible values until a fixed point --- a complete level that satisfies all constraints --- is reached. This process involves iteratively exploring possible assignments of values to variables. The number of possible assignments grows exponentially with the number of variables. Consequently, this iterative process, which may include backtracking, can be very time-consuming.

% therefore we did ...
Recent work in interpretable/explainable AI has led to AI approaches (such as classifiers) that can explain their decisions. The goal of explainability in machine learning is to make the model's decision-making process understandable to humans, allowing users to trust and interpret the model's outputs. For example, in a model designed to recognize handwritten digits, an explainable model might highlight specific regions of the image, such as the curves and lines that form the shape of each digit, indicating that these features significantly contribute to the model's decision that the image contains a particular number. In our work, we propose training a binary classifier to distinguish between two classes of game levels: solvable and unsolvable. We then apply explainability methods to these classifiers to identify the regions of the level that contribute to their unsolvability. Subsequently, we provide meaningful weights that prioritize the change of these problematic regions to a constraint-based solver, guiding level repair and enhancing its performance.

In this work, we first generate a training set consisting of solvable and unsolvable levels from three different 2D tile-based games. These levels are used to train deep neural network classifiers that label levels as either solvable or unsolvable. Explainability methods enable the classifiers to provide per-location attributions, indicating the importance of each tile in the classification process. These attributions assign importance to the features (level tiles) contributing to the classifier's predictions (solvability). We utilize this attribution data as weights to guide a constraint satisfaction solver that repairs the level.

Subsequently, we construct a pipeline for our experiments to measure the performance of this approach. This pipeline includes a level generator, the explainable solvability classifier, a weight generator, and a constraint satisfaction solver. This pipeline generates an unsolvable level, obtains attributions for the unsolvability of each tile location from the explainable classifier, creates weights based on the attributions for the constraint solver, and attempts to repair the level by solving this constraint satisfaction problem, guided by the attributions.

We conducted our experiments on levels across three different 2D tile-based game domains: Super Mario Bros., Super Cat Tales, and a custom Cave game.
% what we found ...
Our results show using attributions to unsolvability gathered from the explainable solvability classifier helps the solver repair the levels faster.

% our contributions are ...
In summary, our contributions are as follows:
\begin{itemize}
  \item We introduce the use of explainability methods to identify regions in unsolvable levels that contribute to their unsolvability.
  \item We develop a pipeline to demonstrate the effectiveness of this approach in enhancing level repair performance.
  \item We present a Mixed Integer Linear Programming encoding of the constraints used for weighted level repair.
\end{itemize}

The GitHub repository containing the data, models, training code, and generated artifacts is available on GitHub\footnote{\url{https://github.com/MahsaBazzaz/ExplainableRepair}}.

%==================================================
\section{Related Work}
\begin{figure*}[t]
  \centering
  \includegraphics[width=1\textwidth]{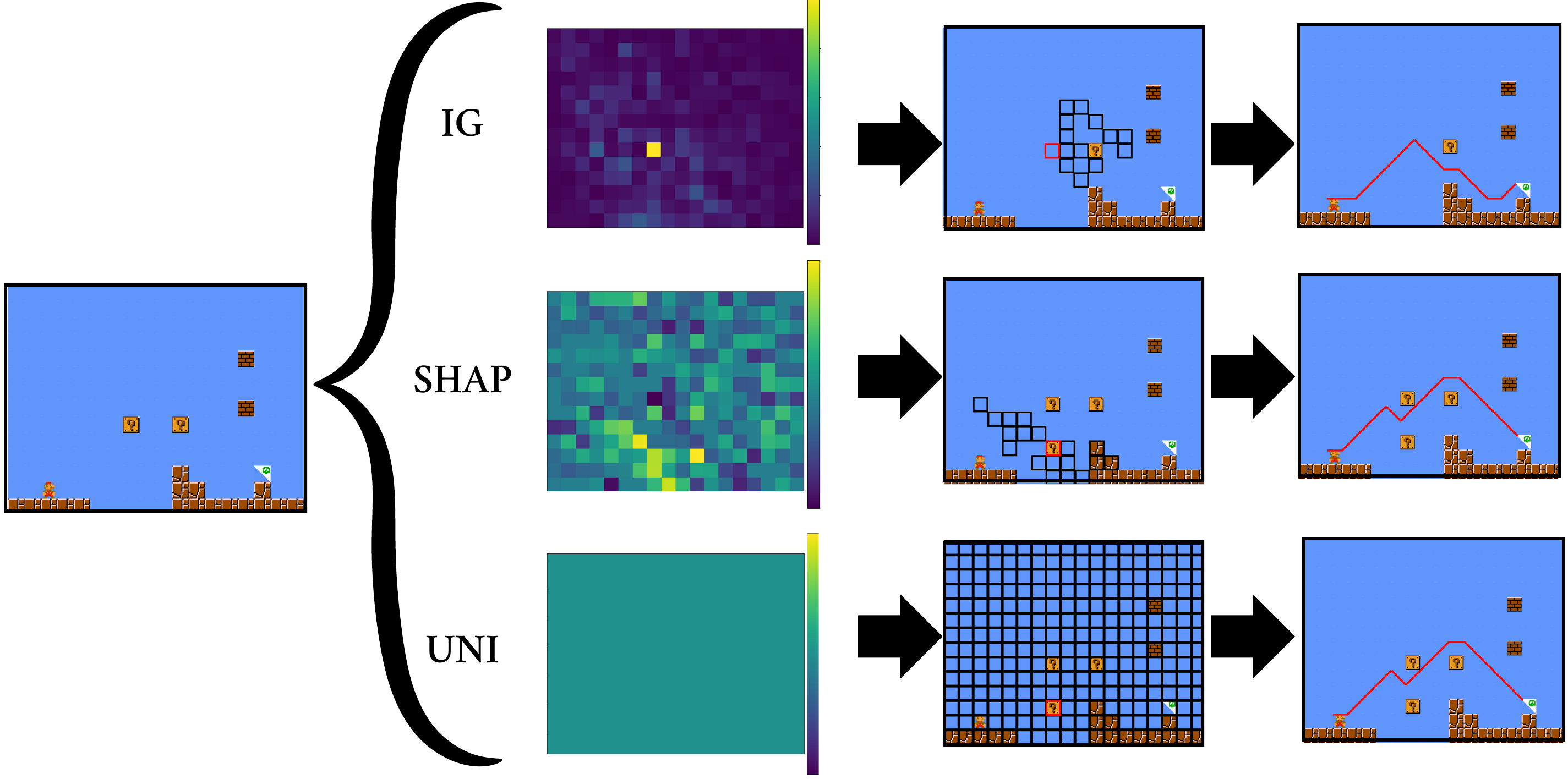}
  \caption{Visual representation of different repairs to a Mario level. The left side is the broken level. Each explainability method generates an attribution for all tiles, and these attributions are scaled to weights. The black squares show the low weights marked for the solver. The red squares show the tile that the solver chooses to change. The right side is the repaired level.}
  \label{fig:mario}
\end{figure*}
\subsection{Level Repair}
Solvability is a crucial aspect of procedural level generation. However, unsolvable levels are almost inevitable since current machine learning models cannot consistently ensure the satisfaction of a constraint like solvability. As a result, various ``generate-then-repair'' approaches have been developed to address this issue. \citet{Cooper2020PathfindingAF} introduced a pathfinding agent capable of repairing levels as part of testing levels for solvability. Their results show that the repair agent could improve the solvability of the level at the cost of an increase in time. \citet{jain2016autoencoders} leveraged the lossy nature of AutoEncoder reproduction to repair broken levels and classify them into different styles. \citet{mott2019controllable} applied long-short term memory recurrent neural network (LSTM) to repair locally incoherent sections of GAN-generated levels. \citet{Zhang2020VideoGL} constructed a framework to repair GAN-generated levels using a mixed-integer linear program (MILP) with solvability constraints. \citet{chen2020image} tried repairing the results of translating image representation (pixels) into a level representation (tiles) with Autoencoders and Markov chains.  Most relevant to our work, \citet{shu2020novel} attempted to repair errors in pipe tile placement using a combination of multi-layer perceptron and genetic algorithms; their approach, similar to ours, trains a model to directly find locations of misplaced tiles. However, in their approach, information about misplaced tiles is used by an evolutionary repairer, and focuses mainly on local tile placements.  On the other hand, our work uses the attributions from explainable classifiers to inform weighted level repair and concentrates primarily on the global solvability of levels.

\subsection{Solvability Analysis}
Path-finding agents and level classifiers have been utilized in order to help generate solvable levels. 
\\ \citet{Snodgrass2014ExperimentsIM}, \citet{volz2018evolving}, and other studies, used Robin Baumgarten’s A* controller \citep{Togelius_Taxonomy_2011} to divide the generated maps into solvable and unsolvable. The tile-based agent of \citet{summerville2015mcmcts} has also been utilized in many projects in order to determine if levels are solvable or unsolvable \citep{summerville2016super,snodgrass2016controllable}.
More recently, deep neural network classifiers have been employed for level solvability classification as well. \citet{bazzaz_active_2023} used Pool-based Active learning to train solvability classifiers that require very few labeled levels.

\subsection{Model Explainability}
Given that recent work has applied deep learning methods to classifying level solvability, explainable classifiers become relevant. Model explainability refers to the degree to which a human is able to understand the reasons behind a choice or prediction made by the model. Explainable machine learning methods can be used to discover knowledge, to debug or justify the model and its predictions, and to improve the model. Machine learning models can be categorized as glass-box or black-box \citep{rai2020explainable}. Glass-box models are inherently explainable to the user, but black-box models need explainability methods to generate explanations. There are multiple techniques used in explainable artificial intelligence (XAI) to understand the decisions made by these complex machine learning models, particularly deep neural networks.

Two widely adopted explainability methods in computer vision are \emph{Deep SHAP} and \emph{Integrated Gradients}, which help in understanding decisions made by classifiers.

\emph{Deep SHAP} is an explainability technique that can be used for models with a neural network architecture \citep{shrikumar2016not}. It combines Shapley values with the DeepLIFT algorithm to approximate the conditional expectations of Shapley values using a selection of background samples \citep{NIPS2017_8a20a862}. Shapley values \citep{shapley1953value} are one of the measures of feature importance from cooperative game theory, to attribute the contribution of each feature to the model's output. It computes the importance of each feature by considering all possible combinations of features and their effects on the prediction.

\emph{Integrated Gradients} provide a robust and interpretable method for understanding the contributions of input features to the predictions of the classifier. It computes the integral of the gradients of the model's output concerning its input along a straight path from a baseline to the input of interest. By integrating the gradients, Integrated Gradients assign importance scores to input features, indicating how much they contribute to the model's decision \citep{sundararajan2017axiomatic}.

\citet{ZhuExplainable} identified Model Explainability as a valuable research area and proposed various research questions and case study scenarios. For example, one suggestion was to make the noise vector in a GAN model explainable, allowing users to control the output by modifying the noise vector.

%%% \todo{cite \cite{gange_optimal_2011} - comparison of SAT and "MIP" solvers for graph layout}
%==================================================
\section{System Overview}
Traditionally in explainable AI, \emph{feature attribution} refers to how much each feature (e.g., pixels of an image) influences the model's output. For example, in a model designed to detect cats in images, feature attribution might highlight specific areas of the image, such as the shape of the ears or the pattern of the fur, indicating that these features significantly contribute to the model's decision that the image contains a cat. This work proposes training a solvability classifier with a dataset of solvable and unsolvable levels. The classifier will then provide relative attributions indicating the contribution of each tile to the classification of a level's solvability. In out work, these attributions can be converted into penalty weights to guide the repair process, where tiles with \emph{higher} attributions to the classification receive \emph{lower} weights and are thus less penalized for changes during the repair. Ideally, this guidance helps the constraint solver to repair the level more efficiently.

To evaluate the effectiveness of weights derived from explainable classification, we developed a controlled pipeline for unsolvable level repair, allowing comparison of various weight generation methods. Figure \ref{fig:system_overview} illustrates the overall framework of the pipeline. The main components of our system are:

\begin{itemize}
\item \emph{level generation} This can be a level generator like constraint-based generators, or any machine learning model like VAEs, GANS, etc.
\item \emph{explainable classification} Takes a level as input and assigns each tile location an \emph{attribution} value of participation in the unsolvability of the level. To train the classifier, a set of \emph{training levels}, both solvable and unsolvable, are needed.
\item \emph{weight generation} Creates meaningful penalty weights for the change of tiles at each location of the level to guide the repair.
\item \emph{level repair} Accepts a level to repair, along with the associated weights, as inputs, and finds a solvable level solution.
\end{itemize}

Here we describe the specific components and domains used in this work.

%--------------------------------------------------
\subsection{Domains}
This work experiments on three different game domains of Super Mario Bros. (Mario)~\citep{GAME_mariobros}, Cave, and Super Cat Tales (Supercat)~\citep{GAME_supercat}.
The Mario levels are based on the level 1-1 from the VGLC \citep{summerville_vglc_2016}.
Cave is a top-down cave map introduced by \citet{cooper2022constraint}; this custom game is created with image tiles from Kenney \citep{WEB_kenney}.
Supercat is also a platforming game with wall and ledge jumps; a simplified version of the actual game’s movement patterns is used.
The example levels as well as patterns and reachability templates used in this work are borrowed from the original work on Sturgeon \citep{cooper2022constraint}. The Mario levels are in size $14\times18$, Cave levels $15\times12$, and Super Cat Tales levels $20\times20$.

%--------------------------------------------------
\subsection{Level Generation}
In this work, we used the Sturgeon constraint-based level generation system \citep{cooper_sturgeon_2022}, which generates levels by converting high-level design rules into constraint satisfaction problems and can use different ``low-level'' solvers to find solutions. This system is also capable of generating unsolvable levels by incorporating additional constraints that a level's goal is not reachable from its start \citep{cooper2024literally}. This constraint-based level generator gave us the controllability to evaluate the performance of weight generation methods regardless of the level generation model.

%--------------------------------------------------
\subsection{Explainable Classification}
\begin{figure}[t]
  \centering
  \includegraphics[width=0.47\textwidth]{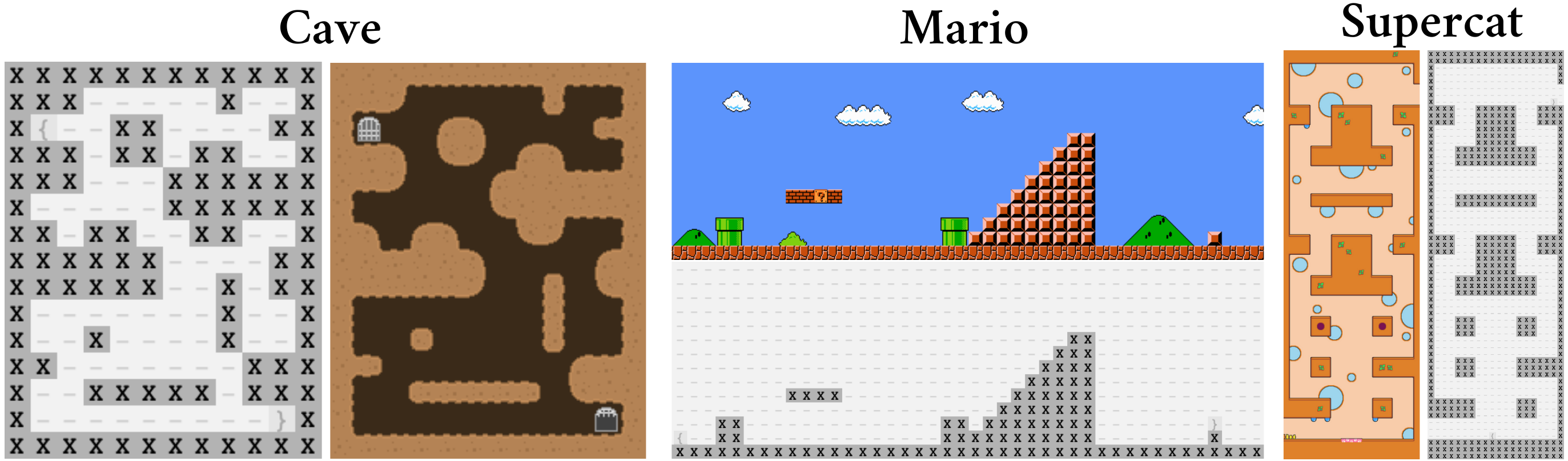}
  \caption{The minimal representation of the levels with 4 tile types (start: \texttt{\{}, goal: \texttt{\}}, solid: \texttt{X}, and empty: \texttt{-}) used for training the classifiers.}
  \label{fig:represantation}
\end{figure}
\begin{figure}[t]
\includegraphics[width=0.47\textwidth]{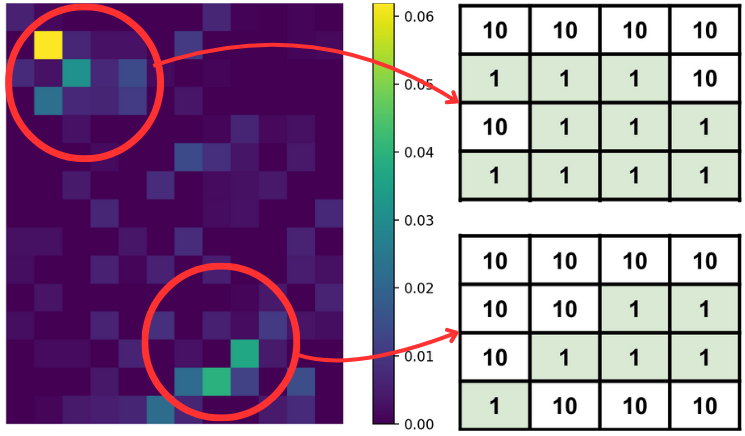}
\caption{Representation of how attributions are scaled to meaningful (positive integers) weights for the solver. Lower values indicated a higher priority to change.}
\centering
\label{fig:attributions_to_weights}
\end{figure}

Figure \ref{fig:attributions_to_weights} illustrates sample attributions of a Mario level and the rescaled values into weights.
We used a unified classifier for the solvability classification of levels in all three domains. We used a minimal representation version of games to train the classifier that consists of 4 tile types. We created a unified simple affordances that include {Solid, Empty, Start, End} and we simplified all games into these affordances. Figure \ref{fig:represantation} shows this new representation for all three domains. After this transformation, the levels are converted into one-hot vectors for the classifier. 

The classifier architecture consists of three fully connected layers followed by a ReLU activation layer to introduce non-linearity, and a Dropout layer to to prevent overfitting. Lastly, the sigmoid activation function is used to squash the output into the range [0, 1] for binary classification. 

To train the classifier, we generated a corpus of $3,000$ unsolvable levels and solvable levels from each domain game. Both solvable levels and unsolvable levels were generated using Sturgeon \citep{cooper_sturgeon_2022}. These levels all maintain the correct game structure, with the only difference being the existence of a path between the start and end points. This makes them ideal for training a classifier to distinguish between solvable and unsolvable levels.

We split the dataset into training and testing sets with an $80-20$ ratio. We used a weight decay of $1e-3$ and a learning rate of $1e-2$ with the Adam optimizer and CrossEntropy loss criterion. The classifier was trained for 10 epochs, achieving an accuracy of $98\%$ for Cave, $97\%$ for Mario, and $100\%$ for Supercat.

We used the implementation of Deep SHAP by the Shap Python package \citep{NIPS2017_7062} and implementation of Integrated Gradients from the Captum Python package \citep{kokhlikyan2020captum} to obtain explanations about tile location attributions in level solvability. These packages are among the most widely used tools for model explainability.

%--------------------------------------------------
\subsection{Weight Generation}
\begin{figure*}[htbp]
    \centering
    \begin{subfigure}[b]{0.325\textwidth}
        \centering
        \includegraphics[width=\textwidth]{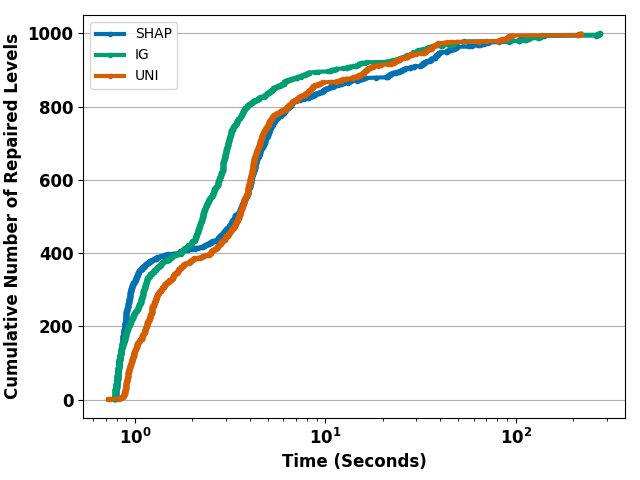}
        \caption{Cave}
        \label{fig:time_cave}
    \end{subfigure}
    \begin{subfigure}[b]{0.325\textwidth}
        \centering
        \includegraphics[width=\textwidth]{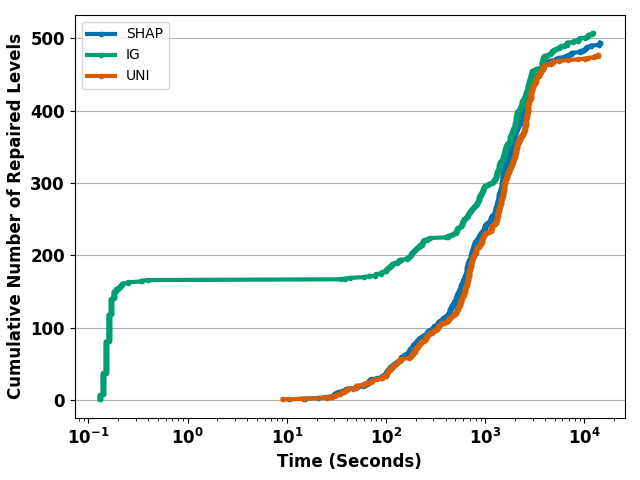}
        \caption{Mario}
        \label{fig:time_mario}
    \end{subfigure}
    \begin{subfigure}[b]{0.325\textwidth}
        \centering
        \includegraphics[width=\textwidth]{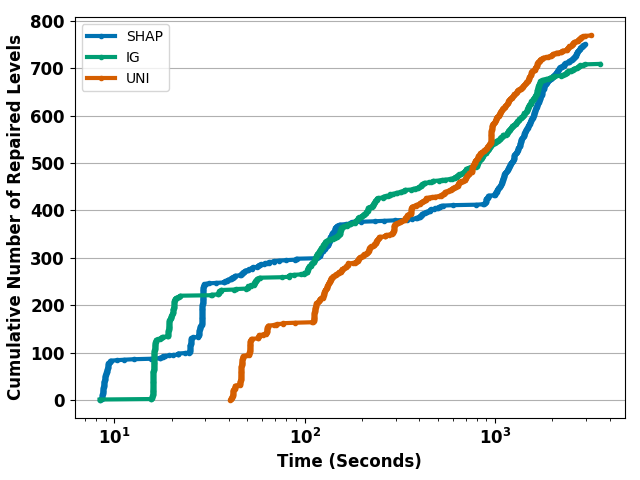}
        \caption{Supercat}
        \label{fig:time_supercat}
    \end{subfigure}
    \caption{Comparison of number of repaired levels in times between different weight generation methods. The time is shown on a logarithmic scale.}
    \label{fig:time_all}
\end{figure*}

\begin{table*}[!h]
    \centering
    \begin{tabularx}{1\textwidth}{|c|XXX|XXX|XXX|}
        \hline
        \multirow{2}{*}{Dataset} & \multicolumn{3}{c|}{Cave} & \multicolumn{3}{c|}{Mario} & \multicolumn{3}{c|}{Supercat} \\
        \cline{2-10}
         & SHAP & IG & UNI & SHAP & IG & UNI & SHAP & IG & UNI \\
        \hline
        Mean & $9.7$ & $7.6$ & $7.8$ & $1192.2$ & $1198.4$ & $1523.2$ & $747.2$ & $511.4$ & $650.5$ \\
        Median & $3.3$ & $2.2$ & $3.5$ & $723.4$ & $722.6$ & $1161.9$ & $148.3$ & $128.5$ & $348.2$ \\
        Std Dev & $25.6$ & $25.3$ & $19.6$ & $1531.9$ & $1533.0$ & $348.2$ & $828.7$ & $688.3$ & $679.6$ \\
        \hline
    \end{tabularx}
    \caption{Statistics of repair times for each domain by each generation method in each game.}
    \label{tab:statistics}
\end{table*}

\begin{table*}[!h]
    \centering
    \begin{tabularx}{1\textwidth}{|c|XXX|XXX|XXX|}
        \hline
        \multirow{2}{*}{Dataset} & \multicolumn{3}{c|}{Cave} & \multicolumn{3}{c|}{Mario} & \multicolumn{3}{c|}{Supercat} \\
        \cline{2-10}
         & SHAP & IG & UNI & SHAP & IG & UNI & SHAP & IG & UNI \\
        \hline
        Mean & $2.9$ & $2.9$ & $2.8$ & $5.1$ & $3.6$ & $1.1$ & $5.2$ & $5.7$ & $4.7$ \\
        Median & $2.0$ & $2.0$ & $2.0$ & $1.0$ & $1.0$ & $1.0$ & $5.0$ & $5.0$ & $5.0$ \\
        Std Dev & $2.0$ & $1.9$ & $1.9$ & $9.5$ & $7.0$ & $2.3$ & $2.6$ & $4.2$ & $0.4$ \\
        \hline
    \end{tabularx}
    \caption{Statistics of the number of changes made by each generation method in each game.}
    \label{tab:changes}
\end{table*}
The tile attributions are processed into weights for the repair solver. The primary goal at this stage is to identify regions of the level with the highest attribution values. To achieve this, the $80^{th}$ percentile threshold of the attributions is calculated. This threshold captures the top $20\%$ of values, effectively filtering out lower values.

Next, a binary map is created by comparing each attribution value against the percentile threshold. Connected component labeling is then employed to extract regions from the original level, focusing exclusively on components that are connected. Using OpenCV's connectedComponentsWithStats function \citep{opencv_library}, we analyze the connected components within the binary map to identify the label corresponding to the largest connected component in terms of area.

Solvers that accept weights as an additional input consider the weights as penalties for changing a tile. Consequently, we assign the largest connected component within the attributions the lowest weight, increasing the likelihood that the solver will modify this area of the level. After identifying the regions with higher attribution values and determining the connected region from the original level, we scale the high values to weight 1 and the others to weight 10. 

%--------------------------------------------------
\subsection{Level Repair}
Sturgeon is capable of repairing unsolvable levels by making minimal changes to the original level \citep{cooper_sturgeon_2022}. This is accomplished by converting the repair task into a constraint satisfaction problem and using constraint solvers to either find a solution or prove that none exists.

Unlike agent-based repair methods, this approach ensures the level is in a definite state after repair. The solver either finds a solution, which may take a considerable amount of time, or confirms that no solution exists, eliminating the need for agents to play the level to verify the repair.

The repair process asks the solver to create a solvable level while imposing soft constraints to match the original level's tile placements as closely as possible. Because this procedure can be more time-consuming than generating a new level, we introduce an enhancement to Sturgeon's original method: the option to assign per-location tile weights. This allows each tile in the level to have a different penalty for deviating from the original layout. These weights are integrated into the constraint problem for the solver.

Essentially, Sturgeon applies a reachability constraint from the start to the end of the level as a hard constraint and uses per-location tile weights as soft constraints to guide the solver in finding a solvable level.

%==================================================
\section{Mixed-Integer Linear Programming Solver}
\begin{figure}[t]
\centering
\includegraphics[width=0.45\textwidth]{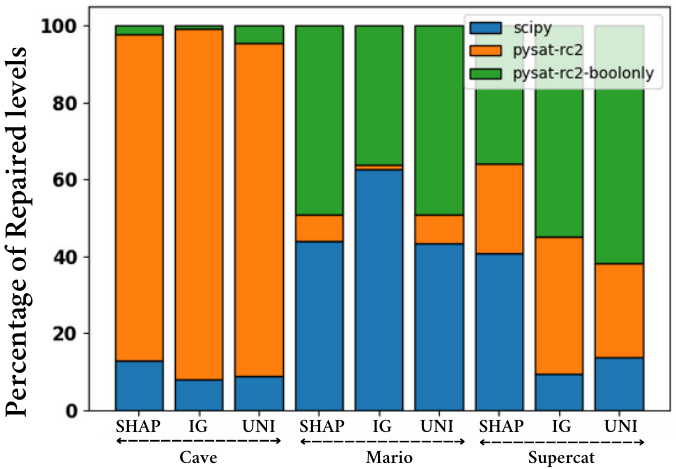}
\caption{Comparison of Percentage of repaired levels by each solver between each weight generation method in each game. Our results show each game has a separate winning solver regardless of the method of weight generation.}
\label{fig:all_solvers}
\end{figure}
Sturgeon uses a small ``mid-level'' API to translate constraint satisfaction problems into ``low-level'' solvers, which do the actual solving \citep{cooper_sturgeon_2022}.  Although previous to this work, Sturgeon had support for several SAT- and ASP-based solvers, in the preliminary development of this project we explored the potential of using Linear Programming approaches for a new low-level solver.  While Linear Programming has been used previously for level repair by \citet{Zhang2020VideoGL}, in their work, they directly encoded the specific problem of level repair, whereas we indirectly encode it through the functions available in Sturgeon's mid-level API in such a way that the encoding could be used for other applications using Sturgeon (e.g. level generation).

\newcommand{\vct}[1]{\mathbf{#1}}

Because Sturgeon's use of a MILP solver is introduced in this work, we provide more information about its implementation. However, the details of the MILP solver are not necessary to follow the other contributions of this work.

Here we describe how Sturgeon formulates its constraint problems for a low-level Mixed-Integer Linear Programming (MILP) solver \citep{scipy_milp_web}. The goal of a MILP is to find assignments for a vector of numerical variables $\vct{x}$ that minimizes the weighted sum of the variables, where the weights are given by another vector $\vct{c}$, i.e. finding $\text{argmin}_{\vct{x}} \vct{c}^\intercal\vct{x}$. The variables can have bounds placed on them directly, i.e. $\vct{s} \le \vct{x} \le \vct{t}$, and bounds can be placed on weighted sums of the variables, i.e. $\vct{u} \le A\vct{x} \le \vct{v}$.  Additionally, given elements of $\vct{x}$ can be required to be integers.

To convert the Boolean constraint satisfaction problem into a MILP, each Boolean variable $v$ in the original problem is given a corresponding numerical variable $x_v$, as well as a corresponding entry in the weight vector $c_v$.  A numerical variable value of 1 corresponds to a Boolean value of true, and a numerical value of 0 corresponds to false.  As Boolean constraints include Boolean \emph{literals} --- both a variable and a \emph{polarity} (i.e. $v$ or $\neg{v}$) --- for a given literal $l$, we also use $x_l$ to refer to the numerical variable for a literal's Boolean variable $x_v$. Note that additional variables can be allocated as needed (described below).

Then, all variables $x_i \in \vct{x}$ have the constraint that they are integers ($x_i \in \mathbb{Z}$) and are either zero or one ($0 \leq x_i \leq 1$), thus $\vct{s} = \vct{0}$ and $\vct{t} = \vct{1}$.  Although this setup could be formulated as a zero-one integer program, in practice we use MILP solvers, which support real-valued variables, but constrain them to be integers.

What remains to complete the MILP is then to fill in the values for the matrix $A$ and the weight vector $\vct{c}$. The weight for all the variables in the initial problem is set to 0 as there is no preference for their being true or false, i.e.
$$c_{v} = 0$$
except for the additional weighting variables described below. To fill in $A$, we specify additional linear constraints on the variables, also described below.

The core of the approach is that, when using the numerical variables to constrain counts of Boolean literals that evaluate to true, positive variables can be included in the constraint directly as $x_v$, and negative variables can be included as $1 - x_v$ with the $1$ then incorporated into the bounds.  For example, a constraint that at most 1 of the literals $i$ and $\neg{j}$ are true would be:

$$x_i + (1 - x_j) \le 1$$

or,

$$x_i - x_j \le 0$$

Thus, if $i$ is true and $j$ is false, both $i$ and $\neg{j}$ will be true, and the constraint will not be satisfied; with other assignments of the $i$ and $j$, it will be satisfied.  For this constraint's row of $A$, the entry for $i$ is set to 1, for $j$ to -1, and the rest to 0; this constraint's entry in $\vct{v}$ is set to 0, and entry in $\vct{u}$ effectively set to $-\infty$ as there is no lower bound.

\newcommand{\polar}{\mathcal{P}}
\newcommand{\negat}{\mathcal{N}}

To construct constraints using larger numbers of literals, we introduce a few utility functions.  These are functions based on the polarity of literals and list of literals:

$$
\setlength{\tabcolsep}{1pt}
\begin{tabular}{ll}
\text{literal polarity:} & $\polar(l) = \begin{cases} x_l & \text{if}~l~\text{is positive} \\ -x_l & \text{if}~l~\text{is negative} \end{cases}$ \\[15pt]
\text{list polarity:} & $\polar(L) = \sum_{l \in L}{\polar(l)}$ \\
\end{tabular}
$$

\noindent and the negativity of literals and list of literals:

$$
\setlength{\tabcolsep}{1pt}
\begin{tabular}{ll}
\text{literal negativity:} & $\negat(l) = \begin{cases} 0 & \text{if}~l~\text{is positive} \\ 1 & \text{if}~l~\text{is negative} \end{cases}$ \\[15pt]
\text{list negativity:} & $\negat(L) = \sum_{l \in L}{\negat(l)}$ \\
\end{tabular}
$$

For example with these functions, the above constraint could be written as:

$$\polar(\langle i, \neg{j} \rangle) \le 1 - \negat(\langle i, \neg{j} \rangle)$$

For Sturgeon's mid-level API, there are several functions to be implemented using a combination of MILP variables (using $\vct{x}$), weights (using $\vct{c}$), and constraints (using $A$).  The most straightforward are: \textsc{MakeVar}(), which simply creates a new numerical variable; \textsc{Solve}(), which calls the MILP solver; and \textsc{GetVar}($v$) and \textsc{GetObjective}(), which access the results of the solver. Our current implementation uses the SciPy MILP solver \citep{virtanen_scipy_2020, huangfu_parallelizing_2018}.

% \newcommand{\posl}[1]{\oplus#1}
% \newcommand{\negl}[1]{\ominus#1}

% We use two utility operators, $\posl{L}$, which gives only the positive literals from a list of literals $L$, and $\negl{L}$, which gives only the negative literals from $L$.

The first mid-level API function that requires additions to the MILP is \noindent\textsc{MakeConj}($L$), which returns a representation of the conjunction (and) of the literals $l \in L$. \textsc{MakeConj} creates a new variable $x_\kappa$ representing the conjunction, and adds a pair of linear constraints such that $x_\kappa$ will be 1 if and only if all $l$ are also 1:

$$
\setlength{\tabcolsep}{1pt}
\begin{tabular}{rcl}
$ |L|x_\kappa - \polar(L)$ & $\leq$ & $\negat(L)$ \\
$-|L|x_\kappa + \polar(L)$ & $\leq$ & $|L| - 1 - \negat(L)$ \\
\end{tabular}
$$

The next mid-level API function is \noindent\textsc{CnstrImpliesDisj}($i, L, w$), which adds a Boolean constraint that the literal $i$ implies the disjunction (or) of the literals $l \in L$.  This function optionally takes a weight $w$.  If no weight is provided, the Boolean constraint is considered hard, and a linear constraint is added:

$$\polar(i) - \polar(L) \le -\negat(i) + \negat(L)$$

\noindent If a weight $w$ is provided, the Boolean constraint is considered soft, and a new weighting variable $x_\omega$ is created and included in the constraint:

$$\polar(i) - \polar(L) - x_\omega \le -\negat(i) + \negat(L)$$
$$c_\omega = w$$

Finally, the \noindent\textsc{CnstrCount}($L, a, b, w$) adds a Boolean constraint that between $a$ and $b$ (inclusive) of the literals $l \in L$ are true.  This function also optionally takes a weight, which, if not provided, treats the Boolean constraint as hard and adds the linear constraint:

$$
a - \negat(L)
\le
\polar(L)
\le
b - \negat(L)
$$

\noindent If a weight $w$ is provided, two new weighting variables $x_\alpha$ and $x_\beta$ are created and included in the constraints:

$$
\setlength{\tabcolsep}{1pt}
\begin{tabular}{rcccl}
$a - \negat(C)$ & $\le$ & $\polar(L) + |L|x_\alpha$ & & \\
                &       & $\polar(L) - |L|x_\beta$  & $\le$ & $b - \negat(L)$ \\
\end{tabular}
$$
% $$
% a - \negat(L)
% \le
% \polar(L) + |L|x_\alpha
% $$
% $$
% \polar(L) - |L|x_\beta
% \le
% b - \negat(L)
% $$
$$c_\alpha = w;~c_\beta = w$$

\noindent Although there are two separate weighting variables, it should not be necessary to set both of them to 1 at the same time.

%==================================================
\section{Experiments}

Here we describe the experiments we ran to evaluate the pipeline and different components. As described above, the pipeline of level generation and level repair uses Sturgeon's impossible level generator to generate unsolvable levels by incorporating constraints that the level's goal is not reachable from the start \citep{cooper2024literally}. The classifier then creates a grid of attribution values of the same size as the level passed into the weight generator, which translates the attributions to a grid of meaningful weights for repair via Sturgeon's weighted constraint solver.
 
The low-level solvers that are included in this study, \tx{pysat-rc2}, \tx{pysat-rc2-boolonly}, and \tx{scipy}, accept positive integer weights. The solvers \tx{pysat-rc2} and \tx{pysat-rc2-boolonly} are Sturgeon's pre-existing PySAT RC2 solver \citep{ignatiev_pysat_2018,ignatiev_rc2_2019}; whereas \tx{pysat-rc2-boolonly} uses only Boolean constraints by encoding cardinality constraints using the PySAT's \tx{kmtotalizer} encoding \citep{morgado_mscg_2014}, \tx{pysat-rc2} uses cardinality constraints directly when possible through MiniCard \citep{liffiton_cardinality_2012}.  The \tx{scipy} solver uses the SciPy MILP solver \citep{virtanen_scipy_2020}, which is itself based on HiGHS \citep{huangfu_parallelizing_2018}. 
The \tx{scipy} MILP solver uses the problem formulation mentioned above.
 
%Lastly,  Sturgeon's level repair has been utilized as a constraint-based solver. This component can be replaced with any other constraint-based solver that accepts weights in input. 

We conducted experiments for $1000$ levels across each domain. The pipeline initially generates a new unsolvable level, and then computes the attributions using Deep SHAP, Integrated Gradients, and a uniform attribution as a baseline. We refer to these three weight generation methods as \textit{SHAP}, \textit{IG}, and \textit{UNI} in our figures and tables to simplify our notation and enhance readability. The three solvers run in parallel for each repair attempt, and the first successful solver terminates the others. We record the winning solver, the time to repair, the time to run the classifier for attributions and weights in the case of \textit{SHAP} and \textit{IG}, and the number of changes made to the original level. Figure \ref{fig:mario} presents the attribution grids generated by each method alongside the corresponding repaired outputs.
% Additional repaired samples are shown in Figure \ref{fig:examples}.

%==================================================
\section{Evaluation}
\begin{figure}[!h]
\centering
\includegraphics[width=0.43\textwidth]{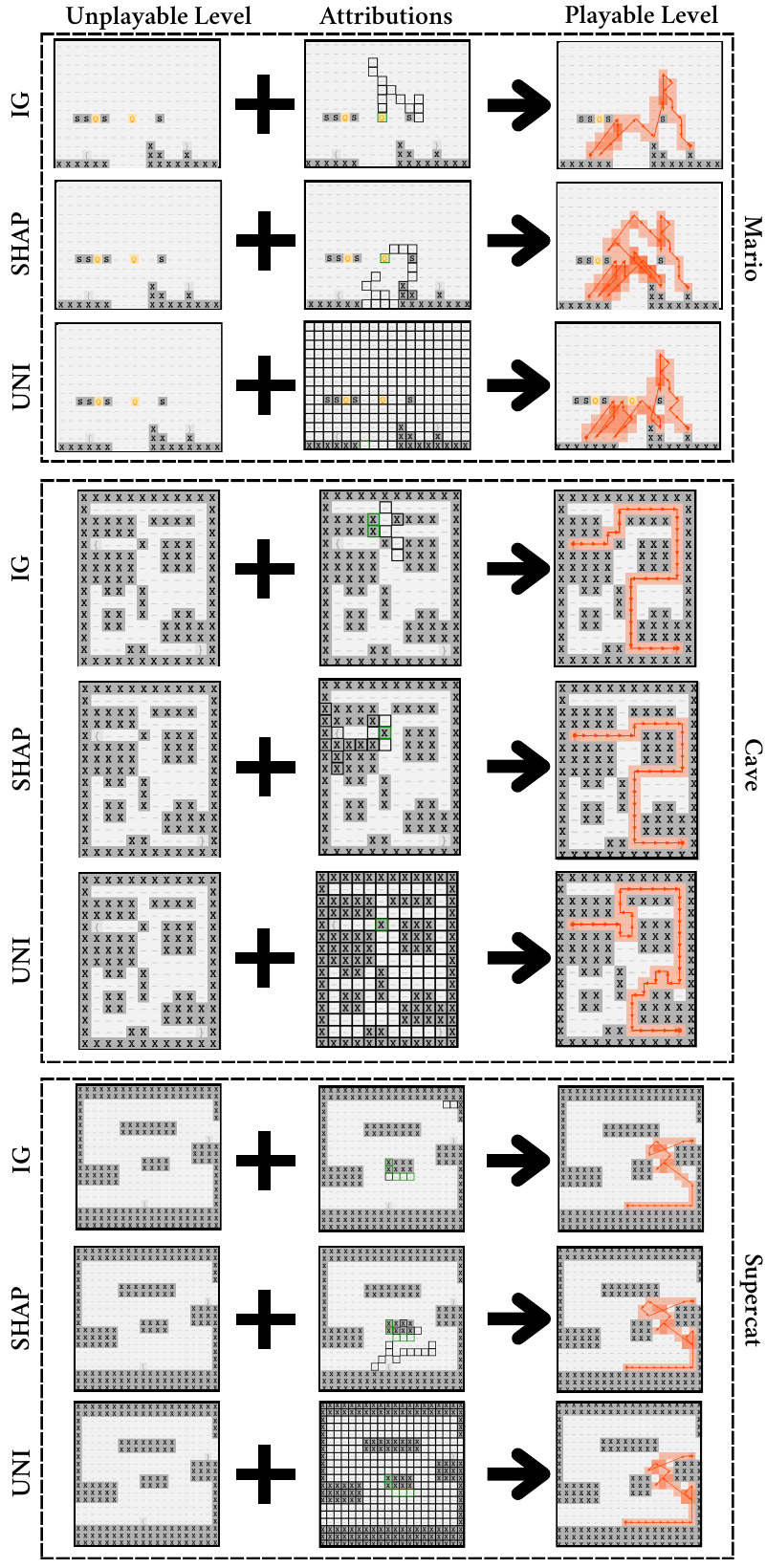}
\caption{Example repaired levels of Mario, Cave, and Supercat. The black squares represent the lowest weights set for repair. The plotted path is just a path between the start and end of the level that the solver came up with during the repair (Not necessarily the shortest path). Sometimes the repaired level is the same between the methods and sometimes it is not.}
\label{fig:examples}
\end{figure} 
We evaluated the performance of this level repair approach based on the duration that it takes to repair a level (including running an explainable classifier if used), the number of changes made to the level to be repaired, and the winning solver that repaired the level. Figure \ref{fig:time_all} shows the cumulative number of repaired levels in time. We used a time limit of $4000$ seconds for each level to be repaired.  
Our results indicate that repair assistance with attributions can accelerate the repair process, particularly for larger and more complex game levels. The Cave levels, being generally easy and quick to repair, did not show as large an improvement with aided repair. On the other hand, Mario and Supercat levels, which are more challenging to repair, showed a reduction in repair time with the aid of attributions. Note that not all the $1000$ has been repaired as some of the repairs were cut off by the time limit. Table \ref{tab:statistics} shows the corresponding detailed statistics of the repair times of different games with different methods, excluding the repair times of levels that were not completed due to the time limit.

We also examined the number of changes made to repair each level to determine if any method requires fewer changes or if any method is prone to making more unnecessary changes. Table \ref{tab:changes} demonstrates no major difference in the number of changes among the methods. Across different games, the median number of changes is consistent, although the means and standard deviations vary. This might be due to the inclusion of different sets of levels, with some methods repairing levels not addressed by others. 

Overall, the results indicate that the repaired levels of all three methods typically involve the same number of tile location changes, although in some cases the attribution-guided repair can result in more tiles being changed. Using attributions as weights for solver results leads to faster repairs without usually affecting the scope of the repair.

Lastly, we investigate which solvers are more successful in repairing the levels with the guide of attributions. As seen in Figure \ref{fig:all_solvers} 
\tx{pysat-rc2} is the best-performing solver for Cave levels, \tx{scipy} excels in Mario levels, and \tx{pysat-rc2-boolonly} performs best in Supercat levels. Interestingly, each solver appears to be particularly well-suited to a specific game. The results of Figure \ref{fig:time_all} and \ref{fig:all_solvers} highlight that this system serves as a guide for constraint solvers to improve the speed of level repair. Thus, the system's effectiveness is inherently tied to the overall performance of the solvers. While levels that these solvers can handle are repaired more quickly, those beyond their capability may remain with no solutions even with additional guidance.

% \begin{figure*}[htbp]
%   \centering
%   \includegraphics[width=1\textwidth]{figs/time.png}
%   \caption{Comparison of number of repaired levels in times between SHAP, IG, and UNIFORM}
%   \label{fig:time_all}
% \end{figure*}

%==================================================
\section{Discussion}

This work focuses exclusively on repairing levels using attribution values derived from an explainable solvability classifier. We see potential in leveraging attribution values from classifiers for other features, such as amusingness, linearity, and more, to generate new levels from existing ones.

It is important to note that to obtain valuable attribution values, it is crucial to use high-quality classifiers. Although training the classifier adds some overhead, this investment pays off by enabling faster level repairs. This approach could be particularly advantageous for larger, more complex segments, where solvers typically encounter the greatest difficulties. In these challenging scenarios, providing the solver with weighted inputs can offer significant benefits. While this project did not address working on larger levels, future research should focus on exploring this area.

For the sake of controllability, this study concentrated on the repair pipeline and employed Sturgeon's impossible level generator to produce unsolvable levels. We acknowledge that using Sturgeon in this pipeline is a current limitation. However, it provided a reliable method for running experiments across a large number of levels and evaluating the performance. We believe this approach can be extended to other level generators that produce undesirable outputs beyond solvability, such as local patterns, structures, and style.

Although this work utilized 2D grid-based levels, we believe the general approach can be applied to other repair contexts where an explainable classifier can attribute features that inform the repair process, such as edges on a mesh or notes in a melody.

%==================================================
\section{Conclusion}
Level repair using constraint solvers is a practice applied in Procedural Content Generation via Machine Learning (PCGML). However, these solvers often require significant time to complete the repair process. To address this issue, we propose the use of explainable solvability classifiers to expedite the repair process. By calculating the contribution of each tile location to the level's unsolvability, we assist the constraint solver with weighted inputs that correspond to these contributions. Our results indicate that this method reduces the repair time while typically changing same number of tiles from the levels.

%\bigskip
%==================================================
%==================================================
\section{Acknowledgments}
Support provided by Research Computing at Northeastern University (\url{https://rc.northeastern.edu/}).
%==================================================

\bibliography{refs}

\end{document}